\documentclass[10pt,twocolumn,letterpaper]{article}

\usepackage{cvpr}
\usepackage{times}
\usepackage{epsfig}
\usepackage{graphicx}
\usepackage{amsmath}
\usepackage{amssymb}
\usepackage{multirow}
\usepackage{diagbox}
\usepackage{mathrsfs}

\usepackage[breaklinks=true,bookmarks=false]{hyperref}

\cvprfinalcopy 

\def\cvprPaperID{****} 

\begin{document}

\title{Distilling a Deep Neural Network into a Takagi-Sugeno-Kang Fuzzy Inference System}

\author{Xiangming Gu\\
Tsinghua University\\
30 Shuangqing Rd, Beijing, China\\
{\tt\small guxm17@mails.tsinghua.edu.cn}
\and
Xiang Cheng\\
National University of Singapore\\
21 Lower Kent Ridge Rd, Singapore\\
{\tt\small elexc@nus.edu.sg}
}

\maketitle

\begin{abstract}
   Deep neural networks (DNNs) demonstrate great success in classification tasks. However, they act as black boxes and we don't know how they make decisions in a particular classification task. To this end, we propose to distill the knowledge from a DNN into a fuzzy inference system (FIS), which is Takagi-Sugeno-Kang (TSK)-type in this paper. The model has the capability to express the knowledge acquired by a DNN based on fuzzy rules, thus explaining a particular decision much easier. Knowledge distillation (KD) is applied to create a TSK-type FIS that generalizes better than one directly from the training data, which is guaranteed through experiments in this paper. To further improve the performances, we modify the baseline method of KD and obtain good results.
\end{abstract}

\section{Introduction}

Since Alex \etal\cite{krizhevsky2012imagenet} adopted deep convolutional neural networks in ImageNet classification and achieved massive success, deep neural network (DNN) became a popular technique and kept the state-of-the-art performance in various applications of artificial intelligence, especially in computer vision. With the advent of VGGNet\cite{simonyan2014very}, GoogleNet\cite{szegedy2015going}, ResNet\cite{he2016deep} and DenseNet\cite{huang2017densely}, the structure of DNN became deeper without loss in generalization. In many cases, DNNs have reached or even exceeded human performance. 

However, it is difficult to apply DNNs in several fields, such as medicine or autonomous cars since interpretability is extremely important in those applications where the reliance of the model must be guaranteed while DNNs are not an interpretable models. Their excellence depends on abstract representations in their hidden layer which are hard to understand and how they make decisions is still unknown. To tackle the problem, techniques for interpreting and understanding DNNs have emerged as an important topic. 

 One solution is network visualization. Researchers aim to visualize the input regions or the feature maps contributing to the final decisons. Among them, Ramprasaath \etal\cite{DBLP:journals/corr/SelvarajuDVCPB16} proposed Gradient-weighted Class Activation Mapping (Grad-CAM) to produce a coarse localization map highlighting the important regions in the image for predicting the categories while Zeiler \etal\cite{zeiler2014visualizing} introduced a Deconvolutional Network (deconvnet) to visualize and understand convolutional neural network by displaying the features. 
 To determine how relevant input or the features for explaining the output of neural network, Layer-wise relevance propagation (LRP) is proposed\cite{Bach2015OnPE}, which is a backward propagation technique.
 Instead of fixing the input and observing the response, Montavon\etal\cite{montavon2018methods} summarized a method of activation maximization to search for an input pattern that produces a maximum model response for a quantity of interest. Network diagnosis is another direction of explainable deep learning. Among the researches, network dissection\cite{bau2017network} is presented to quantify the interpretability of latent representations of CNNs by evaluating the alignment between individual hidden units and a set of semantic concepts. And influence function\cite{koh2017understanding} is proposed to understand model behaviors, debug models, detect dataset, errors and even create visually-indistinguishable training-set attacks.
 
 In this paper, we focus on explainable models instead of analyzing DNNs. Several trials have been done, e.g., Explainable Neural Network is proposed based on additive index models and adopts multi-layer perceptrons to learn the ridge functions\cite{vaughan2018explainable}. To promote this work, adaptive explainable neural network is designed for achieving both the good predictive performance and model interpretability\cite{chen2020adaptive}. However, neural networks still play important roles in the framework and it is hard to understand how they approach the ridge functions. To circumvent directly applying DNNs in tasks, Frosst\etal\cite{frosst2017distilling} proposed to distill a neural network into a soft decision tree and then apply this model to make hierarchical decisions based on the advent of knowledge distillation\cite{hinton2015distilling}. Inspired by this work, Gradient Boosting Trees\cite{che2016interpretable} and global additive model\cite{tan2018learning} are learned through knowledge distillation to improve the generalization and interpretability. 

We adopt another interpretable model: Takagi-Sugeno-Kang (TSK)-type fuzzy inference system (FIS)\cite{takagi1985fuzzy, cui2020optimize}, which is initially designed to learn human knowledge. Therefore, it is natural to adopt machine knowledge, which is more powerful than human experts. To our knowledge, there are limited researches about adopting knowledge distillation on FIS. Hence, we propose to distill knowledge from DNNs into TSK-type FIS and  give an explanation on how TSK-FIS makes decisions through rule-based framework. To build more comprehensible fuzzy rules, we extract features of input through PCA. Then we pre-train teacher model on raw data while the training student model online on features inspired by cross-modality knowledge distillation\cite{gou2020knowledge}. In addition, We also propose a modified method of KD considering the difference of generalization between teacher and student model.
 

\section{Methodology}

In this section, we first evaluate the mathematical expressions of TSK-type FIS. Then the baseline method of knowledge distillation is described. At last, we demonstrate the framework of proposed novel method. In addition, we describe a modified version of knowledge distillation, which further improves the effectiveness of knowledge distillation on TSK-type FIS.

\subsection{TSK-type FIS}
Given a labeled training set with n samples $\left\{(X^1,l^1),(X^2,l^2),...,(X^n,l^n) \right\}$ and $l^i\in\left\{0,1,2,...,C-1 \right\}$, where C stands for the number of categories. $X^i$ is a vector, which represents features extracted or processed from a raw training sample while $l^i$ corresponds to its label.

Suppose TSK-type FIS has R rules in the following form:
\begin{equation}
    \text{Rule}_r\text{: If}\, x_1\, \text{is}\, A_{r,1}\, \text{and ... and}\, x_k\, \text{is}\, A_{r,k}, \,\text{then}\, Y=AX+b
\end{equation}

Where $A_{r,d}, r\in{1,2,...,R},d\in{1,2,...,D}$ stands for the fuzzy set of d-th premise part in the r-th rule and Y refers to a C-dimensional vector of which each dimension stands for score assigned to each category. The membership of each dimension is measured by a Gaussian function whose center $c_{r,d}$ and variance $\sigma_{r,d}$ are trainable parameters:
\begin{equation}
    \mu_{A_{r,d}}(x_d)=\exp{(-\frac{(x_d-c_{r,d})^2}{2\sigma_{r,d}^2})}
\end{equation}

Then the degree of fulfillment of each rule is calculated as:
\begin{equation}
   f_r(X)=\cap_{d=1}^D \mu_{A_{r,d}}(x_d)
\end{equation}

Where $\cap$ is a T-norm operator, which can be the minimum or product value of all elements. In this paper, we use product operator to compute the degree of fulfillment. After that, perform normalization to obtain the weight for each rule.
\begin{equation}
    \overline{f_r}(X)= \frac{f_r(X)}{\sum_{i=1}^{R} f_i(X)}
\end{equation}

The consequence of TSK-type FIS for each rule is:
\begin{equation}
    Y_r=A_r X+b_r
\end{equation}

Where $A_r$ is a matrix with the size of $C\times D$ and $b_r$ is a C-dimensional vector. These parameters are determined by training process.

The $\textsl logits$ are the weighted average on consequences of all rules:
\begin{equation}
    Y=\sum_{r=1}^{R} \overline{f_r}(X)Y_r
\end{equation}

Finally, we apply a softmax layer accepting the logits to obtain the probability distribution.
\begin{equation}
    P=\text{softmax}(Y)
\end{equation}

\begin{figure}[t]
\begin{center}
\includegraphics[width=0.6\linewidth]{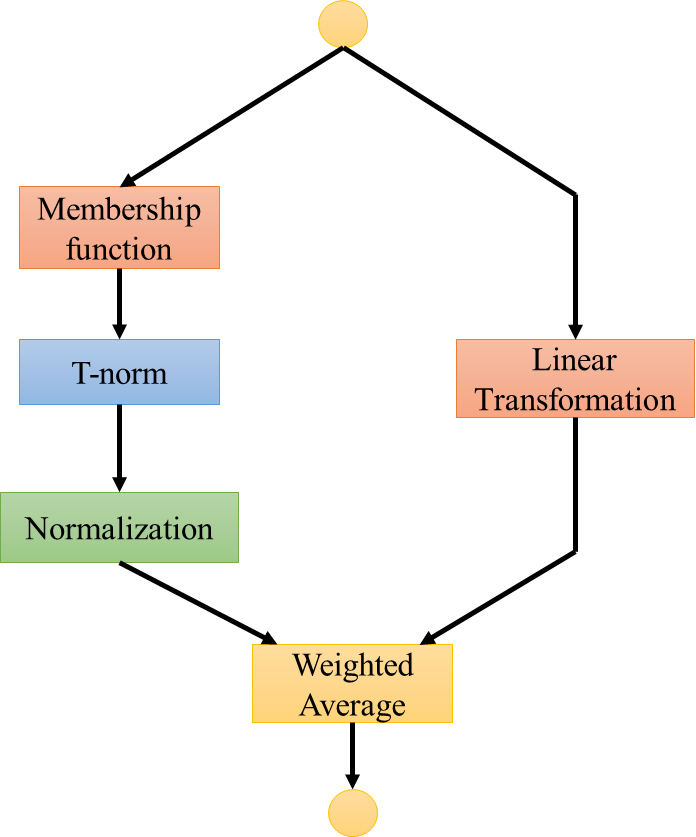}
\end{center}
   \caption{Architecture of TSK-type FIS.}
\label{fig1}
\end{figure}

\subsection{Knowledge distillation}
In this paper, we firstly apply the baseline algorithm of knowledge distillation\cite{hinton2015distilling}, which transfers the response-based knowledge from the teacher model. Normally, the output of a neural network for classification is a probability distribution over categories, which is created by adding a softmax layer over the output of the last fully connected layer, also known as $\textsl logits$. Suppose logits of teacher model are $Y^{(t)}=\text{NN}(X)$ and the logits of FIS are $Y^{(s)}=\text{FIS}(X)$. By introducing a parameter named as temperature $T$, the generalized softmax layer converts logits into softened probability distribution:
\begin{equation}
    Q=\text{softmax}(\frac{Y}{T})
\end{equation}

Normally, $T$ is set to 1. However, using a higher value of $T$ produces a more softened probability distribution, which reveals the similarity of different categories. $q^{(t)}$ is also called soft labels. Compared with hard labels, referring to one-hot labels, soft labels contain the dark knowledge acquired by the pre-trained teacher model.

To learn the response-based knowledge from teacher model, Hinton \etal\cite{hinton2015distilling}. proposed to minimize KL divergence between the softened probability distribution of student model and soft labels.
\begin{equation}
    L_\text{KD}=\frac{1}{N}\sum_{i=1}^{N}(-\sum_{j=1}^{C} q_{i,j}^{(t)}\log q_{i,j}^{(s)})=\frac{1}{N}\sum_{i=1}^{N}\text{KL}(Q_i^{(t)}, Q_i^{(s)})
\end{equation}

Since hard labels have been provided, the commonly used cross-entropy loss for pure supervised learning is represented as:
\begin{equation}
    L_\text{S}=\frac{1}{N}\sum_{i=1}^{N}(-\log p_{i, l_i}^{(s)})=\frac{1}{N}\sum_{i=1}^{N}\mathcal{H}(P_i, l_i)
\end{equation}

where $R$ refers to the one-hot real label for each sample and $P$ refers to the output of student model when $T$ equals to 1.

Finally, the loss function for KD is a weighted sum of loss $L_{\text{KD}}$ and $L_{\text{S}}$. The weight $\alpha$ is a hyper-parameter, which is determined before training the student model.
\begin{equation}
    L=(1-\alpha)L_\text{S}+\alpha T^2L_{\text{KD}}
\end{equation}

\subsection{Distill knowledge from DNN into TSK-type FIS}

\begin{figure}[t]
\begin{center}
\includegraphics[width=1.0\linewidth]{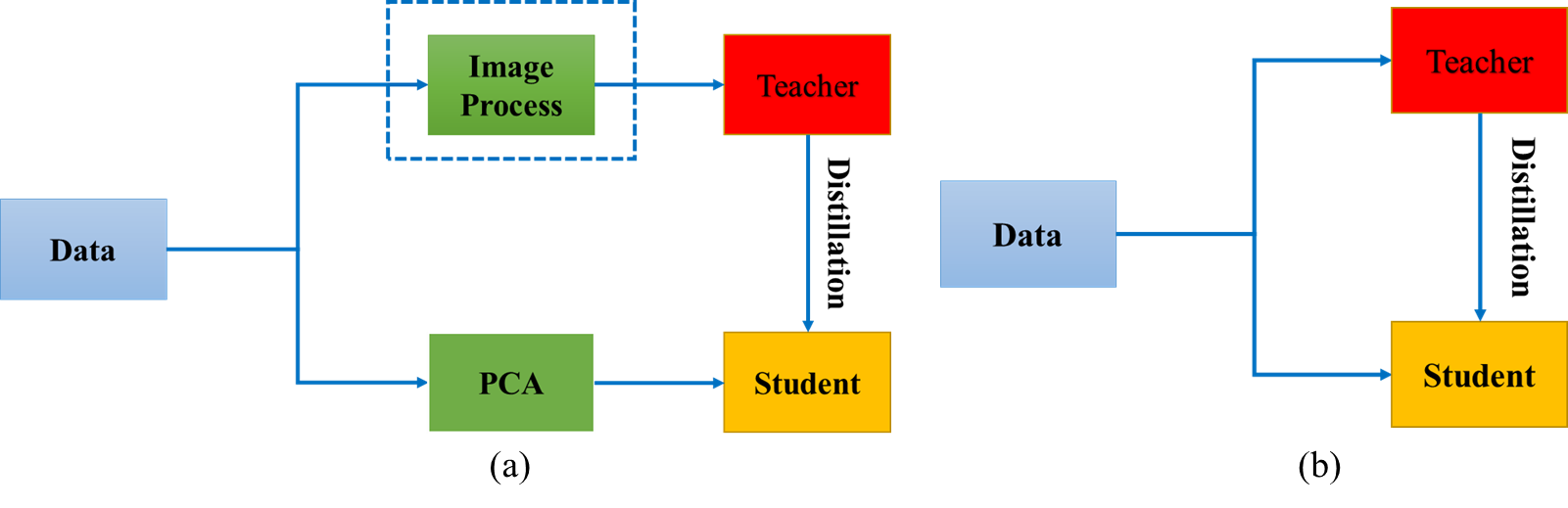}
\end{center}
   \caption{The framework of KD from a DNN into a TSK-type FIS. Teacher model is pre-trained while student model is trained online. (a) For image classification, teacher model is trained on raw data or the data after image processing. Student model is trained on the features extracted by PCA. (b) When the data is low dimensional, both the teacher model and student model are trained on raw data.}
\label{fig2}
\end{figure}

The main idea of training TSK-type FIS from neural network is depicted in Figure \ref{fig2}. When the dimension of input is low, we can directly train the student model from raw data with normalization. However, when the input are high-dimensional images, the generalization of TSK-type FIS will be limited and the training process will be also slowed down. To overcome the obstacles, we perform PCA to reduce the dimension of input and then train the student model. But for teacher model, training process proceeds directly on images. Therefore, to improve its performances, several image processing algorithms can be performed. For instance, we can normalize the pixels, randomly crop images, horizontally flip the images with a given probability, and randomly rotate the images. 

When directly training a model, we apply cross-entropy loss function expressed as (10). Therefore, the probability distribution output by the model will mimic the hard labels, which means that the probability of true category will be close to one and probability of other categories will approximately equal to zero. Since the generalization of student model is normally less than teacher model, its probability distribution will be more 'softened' than teacher model. However, the difference isn't considered in the baseline KD. 

We modify the baseline method of KD based on the introduction of two unequal temperatures. In this way, the probability distributions of teacher model and student model are softened to different degrees. To simplify, we set $T_2=1$ in this paper.
\begin{equation}
    Q^{(t)}=\text{softmax}(\frac{Y}{T_1})
\end{equation}
\begin{equation}
    Q^{(s)}=\text{softmax}(\frac{Y}{T_2})
\end{equation}

Therefore, combined equation (9) with equation (10), the loss function will be expressed as:
\begin{equation}
    L_{\text{new}}=(1-\alpha)L_\text{S}+\alpha T_1T_2L_{\text{KD}}
\end{equation}
\section{Experiments and Results}
 In this section we first provide the implementation details and experimental settings. We then investigate and compare the effects of baseline KD and modified KD on TSK-type FIS. Finally, we illustrate the decision-making process of TSK-type FIS.
 
\subsection{Experimental Settings}
We conduct our experiments on two well-known image classification tasks: MNIST and FashionMNIST. Both datasets comprise of 60K training images and 10K validation images annotated with labels from a set of 10 classes. The images are all $28\times28$.

\begin{figure}[t]
\begin{center}
\includegraphics[width=0.8\linewidth]{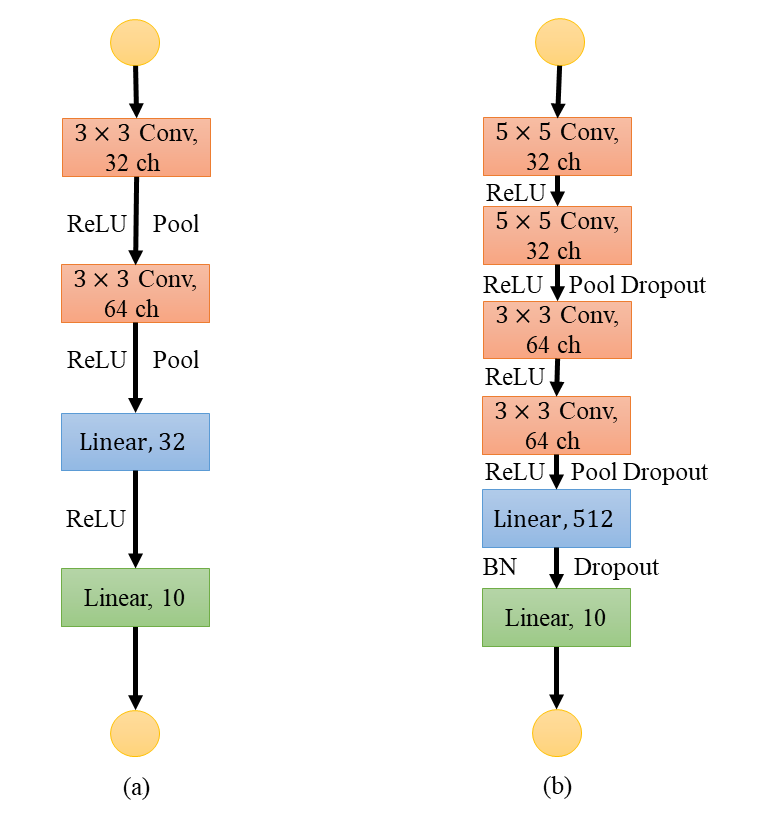}
\end{center}
   \caption{Architecture of teacher model (a) CNN for MNIST (b) CNN for FashionMNIST.}
\label{fig3}
\end{figure}

We select convolutional neural networks (CNNs) as teacher models whose architectures are displayed in Figure\ref{fig3}. Since FashionMNIST is more difficult than MNIST, we apply batch normalization\cite{ioffe2015batch} and dropout layer\cite{10.5555/2627435.2670313} designing the teacher model. We perform no data augmentation. During the training process, we apply Adam\cite{kingma2014adam} as the optimizer , cross-entropy as the loss function and set the mini-batch size 64. After training, the accuracy of teacher model on MNIST validation set is 99.10\% while the accuracy on FashionMNIST validation set is 93.15\%.

To train the student models, we firstly perform PCA on flattened images and extract 64-dimensional features. Then we select the number of fuzzy rules from $\left\{3,4,5,...,14,15\right\}$ and initialize the center and variance of each rule by fuzzy c-means clustering implemented on Sklearn. During the training, we still apply Adam as optimizer, set the mini-batch size as 64 and initial learning rate as 0.01. The student models are trained for 100 epochs with the learning rate divided by 2 for each 25 epochs.

We apply grid search to determine the best hyper-parameters for KD. Temperature is chosen from $\left\{1, 2.5, 5, 7.5\right\}$ and $\alpha$ is selected from $\left\{0.25, 0.5, 0.75, 1\right\}$. To demonstrate the effectiveness of KD, other hyper-parameters are the same as that without KD.

\subsection{Effects of knowledge distillation}

\begin{figure}[t]
\begin{center}
\includegraphics[width=1.0\linewidth]{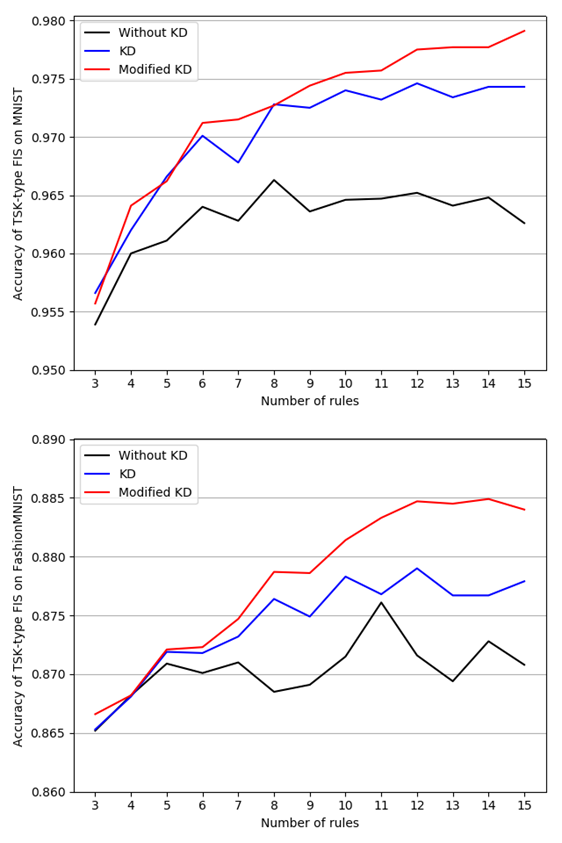}
\end{center}
   \caption{Comparison results of proposed methods. \textbf{Top}: Validation accuracy on MNIST dataset. \textbf{Bottom}: Validation accuracy on FashionMNIST dataset.}
\label{fig4}
\end{figure}

The classification accuracy on validation set is illustrated in Figure \ref{fig5}. It is noticeable that the performances of TSK-type FIS have improved to some extent with increase of the number of rules. Without knowledge distillation, the best accuracy of student model on MNIST is 96.63\% while the accuracy on FashionMNIST is 87.61\%. After the baseline KD, the performances reach 97.46\% and 87.83\% separately. When we modify the baseline KD, the results are much better$-$97.91\% on MNIST and 88.49\% on FashionMNIST.

According to the curve, KD promisingly improves the performances of student models and enables them to be applied in classification tasks with requirements for explanation. Furthermore, it is clear that modified KD is more promising. Under the same number of fuzzy rules, the maximum improvements of baseline KD on student model are 1.2\% and 0.8\% for MNIST and FashionMNIST separately while modified KD reaches 1.6\% and 1.5\%.

Apart from difference in capability to explain the decision-making process, we also compare parameters and CPU time of teacher model and student model with 15 fuzzy rules, which are displayed in Table \ref{table1}. It is clear that TSK-type FIS has much less trainable parameters than CNN model, which means it takes up smaller storage. When it concerns CPU time, which is calculated by separately training teacher and student model for 10 epochs on CPU and then dividing the running time by 10, we notice that student model is much easier to train than teacher model.

\begin{table}
\begin{center}
\begin{tabular}{c|c|c|c|c|c}
\hline
\multicolumn{2}{c|}{\multirow{2}{*}{Dataset}} & \multicolumn{2}{c|}{Parameters} & \multicolumn{2}{c}{CPU time}\\
\cline{3-6}
\multicolumn{2}{c|}{} & Teacher & Student & Teacher & Student\\
\hline\hline
\multicolumn{2}{c|}{MNIST} & 120K & \textbf{12K} & 53s & \textbf{3.3s}\\
\multicolumn{2}{c|}{FashionMNIST} & 1694K & \textbf{12K} & 167s & \textbf{3.3s}\\
\hline
\end{tabular}
\end{center}
\caption{Comparison of teacher and student model.}
\label{table1}
\end{table}

We also compare our results with other related methods. Decision trees (DTs) with knowledge distillation\cite{frosst2017distilling} reaches the accuracy of 96.76\%, which is 1.2\% lower than TSK-type FIS. The difference proves that TSK-type FIS has higher generalization than decision tree as explainable models. In addition, Zhen \etal\cite{jardim2020intelligent} proposed an interpretable convolution neural networks using rule-based framework. The main idea of the algorithm is to combine CNN and TSK-type FIS, and then train them as a whole model. The CNN part is regarded as a backbone network to extract the features, while TSK-type FIS is considered as a classifier. This design adopts image representations in the hidden layer of CNN, which are difficult to interpret. It is noticeable that without those representations, we still obtain satisfying results, which vindicates the sound effects of KD.

\begin{table}
\begin{center}
\begin{tabular}{c|c|c}
\hline
Dataset & MNIST & FashionMNIST \\
\hline\hline
DT+KD\cite{frosst2017distilling} & 96.76\% & -  \\
CNN+TSK\cite{jardim2020intelligent} & 97.52\% & 84.50\% \\
Ours & \textbf{97.91}\% & \textbf{88.49}\% \\
\hline
\end{tabular}
\end{center}
\caption{Comparisons of different models on benchmark datasets.}
\label{table2}
\end{table}

\subsection{Explanation of decision-making process}
\begin{figure*}[t]
\begin{center}
\includegraphics[width=1.0\linewidth]{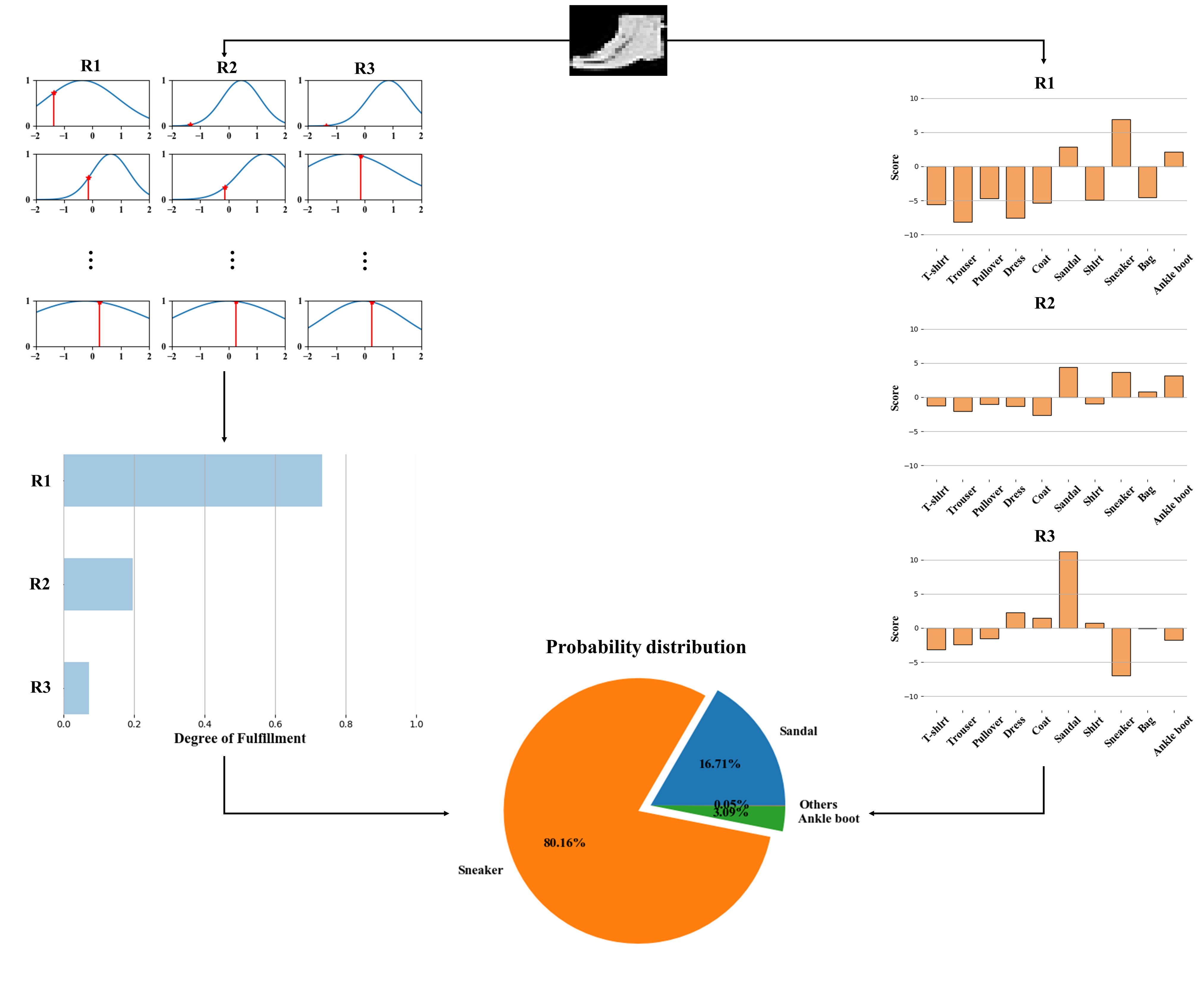}
\end{center}
   \caption{Explanation of TSK-type FIS on decision-making.}
\label{fig5}
\end{figure*}

The decision-making process of TSK-type FIS is displayed in Figure \ref{fig5}. Accepting an image of sneaker as input, we firstly flatten all the pixels and then perform PCA to reduce its dimension to 16. Then we separately calculate the premise part and consequent part. In the premise part, we have trained 3 rules, each of which is represented by 16 membership functions. The height of red vertical lines represent the membership value of each dimension in an individual rule. It is simple to decide which dimension of features makes the most contribution to the degree of fulfillment. Performing the T-norm operator and normalization, we obtain the degree of fulfillment for each rule. In this case, the rule 1 is fired most.

 About the consequent part, we perform a linear transform to obtain scores over all categories for each rule. We notice that for rule 1, the score of sneaker which is the real category is the highest, which explains the firing strength of rule 1 is the largest. What deserves attention is that the scores of sneak, sandal and ankle boot are the highest among all categories in rule 1 and rule 2, which reflects the similarity among these three categories of shoes. In addition, the score of sneaker is remarkable low in rule 3 since fulfillment for this rule is small. 
 
Finally we average the scores applying the degree of fulfillment as weights and add a softmax function to transform logits into probability distribution as shown in pie graph. The classification probability of sneaker is the highest, which indicates the correctness of decisions making by TSK-type FIS. Since sneaker is similar to ankle boot and sandal, it is not surprising that these two categories have higher probability than others.

\section{Conclusion and Future work}
In this paper, we present insights in distilling knowledge from DNN into TSK-type FIS and explanation about how FIS makes decisions in a given classification task. Through experiments, we demonstrate higher performances of this method compared to decision tree with knowledge distillation and a combined model of CNN and TSK-type FIS. We also propose a modified method of KD and prove its effectiveness.

In the future, we will adopt this algorithm in other applications, like control, speech recognition, NLP etc. To further improve the performances of student model, adversarial distillation\cite{xu2017training} and other algorithms of knowledge distillation can be considered. When performing other algorithms, two unequal temperatures may be introduced to improve the performances.  

{\small
\bibliographystyle{ieee_fullname}
\bibliography{egbib}
}

\end{document}